\documentclass[twoside,leqno,twocolumn]{article}
\usepackage{ltexpprt}
\usepackage{comment}
\usepackage{amsmath}
\usepackage{amsfonts}
\usepackage{newlfont}
\usepackage{graphicx}
\usepackage{epstopdf}
\usepackage{subcaption}
\usepackage[strings]{underscore}
\usepackage{pdfpages}
\usepackage{xr}
\begin{document}

\title{\Large Statistical Learning Theory Approach for Data Classification with
$\ell$-diversity
\thanks{Supported by the Northrop Grumman Cybersecurity Research Consortium}}
\author{Koray Mancuhan \\
\and
Chris Clifton\thanks{Purdue University-West Lafayette; Department of Computer Science and CERIAS}
\thanks{\{kmancuha,clifton\}@cs.purdue.edu}}
\date{October 15 2016}
\maketitle


\begin{abstract} \small\baselineskip=9pt 
Corporations are retaining ever-larger corpuses of personal data; 
the frequency or breaches and corresponding privacy impact have 
been rising accordingly. One way to mitigate this risk is through use 
of anonymized data, limiting the exposure of individual data to only 
where it is absolutely needed. This would seem particularly appropriate 
for data mining, where the goal is generalizable knowledge rather than 
data on specific individuals. In practice, corporate data miners often 
insist on original data, for fear that they might "miss something" with 
anonymized or differentially private approaches.
This paper provides a theoretical justification for the use of anonymized data.
Specifically, we show that a support vector classifier trained on anatomized 
data satisfying $\ell$-diversity should be expected to do as well as on the original data.
Anatomy preserves all data values, but introduces uncertainty in the 
mapping between identifying and sensitive values, thus satisfying $\ell$-diversity.
The theoretical effectiveness of the proposed approach is
validated using several publicly available datasets, showing that
we outperform the state of the art for support vector classification using 
training data protected by $k$-anonymity, and are comparable to 
learning on the original data.
\end{abstract}

\section{Introduction}

Many privacy definitions have been proposed
based on generalizing/suppressing data ($\ell$-diversity\cite{ldiversity}, 
$k$-anonymity \cite{kanon_defn,kanon_sweeney}, $t$-closeness 
\cite{tcloseness}, $\delta$-presence \cite{NergizDpresence}, 
($\alpha$,$k$)-anonymity \cite{EnhancedKanon}). 
Other alternatives include value swapping \cite{CensusRelease},
distortion \cite{AgrawalPpdm}, randomization \cite{EvfimievskiBreach}, 
and noise addition (e.g., differential privacy \cite{DifferentialPrivacy}). 
Generalization consists of replacing identifying attribute values with 
a less specific version \cite{kanon_sweeney}. Suppression can be 
viewed as the ultimate generalization, replacing the identifying value 
with an ``any'' value \cite{kanon_sweeney}. Generalization has
the advantage of preserving truth, but a less specific truth that 
reduces utility of the published data.

Xiao and Tao proposed anatomization as a method to enforce 
$\ell$-diversity while preserving specific data values \cite{XiaoAnatomy}. 
Anatomization splits instances across two tables, one containing 
identifying information and the other containing private information.
The more general approach of fragmentation \cite{fragmentation}
divides a given dataset's attributes into two sets of attributes 
(2 partitions) such that an encryption mechanism avoids associations 
between two different small partitions. Vimercati et al. extend 
fragmentation to multiple partitions \cite{loose_fragmentation},
and Tamas et al. propose an extension that deals with multiple 
sensitive attributes \cite{l_diver_multiple}. The main advantage of
 anatomization/fragmentation is that it preserves the original
values of data; the uncertainty is only in the mapping between 
individuals and sensitive values.

We show that this additional information has real value.
First, we demonstrate that in theory, learning from anatomized data
can be as good as learning from the raw data.  We then demonstrate
empirically that learning from anatomized data beats learning from
generalization-based anonymization.

\emph{This paper looks at linear support vector (SVC) and support vector 
machine (SVM) classifiers}. This focus was chosen because these classifiers 
have a wide range of successful applications,
and also have some solid theoretical basis
their generalization properties. We propose a simple heuristic to 
preprocess the anatomized data such that SVC and SVM generalize well with 
sufficiently large training data.

There is concern that anatomization is vulnerable to several attacks 
\cite{Definetti,AnatomyImprov,NinghuiSlicing}. While this can be an issue, 
\emph{any} method that provides meaningful utility fails to provide perfect 
privacy against a sufficiently strong adversary
\cite{NinghuiTradeoff,DifferentialPrivacy}.
Introducing uncertainty into the anonymization process reduces the risk of 
many attacks, e.g., minimality \cite{MinimalityAttack,CormodeMinimality}.
Our theoretical analysis holds for any assignment of items to anatomy groups, 
including a random assignment, which provides a high degree of robustness against
minimality and correlation-based attacks.
While this does not eliminate privacy risk, if the alternative is to use
the original data, we show that anatomy provides comparable utility while
reducing the privacy risk.
This paper has the following key contributions: 

\begin{enumerate}
	\item We define a classification task on anatomized data without violating the
	random worlds assumption. A violating classification task would be the
	prediction of sensitive attribute, a task that was found to be \#P-complete 
	by Kifer \cite{Definetti}.
	
	\item We propose a heuristic algorithm to train SVC and SVM when 
	the test data is neither anonymized nor anatomized. Inan et al. already 
	gives a practical applications of such a learning scenario \cite{kAnonSvm}.
	
	\item We study the effect of our heuristic algorithm on the generalization
	error. To our best knowledge, this is the first paper 
	in the privacy community that does such analysis for $\ell$-diversity
	
	\item In empirical analysis, our algorithm will be compared with SVM and SVC
	that are trained on either unprotected data or generalized
	data (under $k$-anonymity \cite{kAnonSvm}). The analysis will be justified with
	the statistical learning theory \cite{Vapnik98, Burges98}
\end{enumerate}

We next summarize related work and define the problem statement. We then give 
necessary definitions and notations. Section \ref{sec:algo} proposes the heuristic 
algorithm and gives theoretical analysis. Empirical analysis is presented in section \ref{sec:exps}. 
Section \ref{sec:conc} summarizes the work and gives future directions.

\section{Related Work and Problem Statement}

There have been studies of linear classification for anonymized data. 
Agrawal et al. proposed an iterative distribution reconstruction algorithm for
distorted training data from which a C4.5 decision tree classifier was trained \cite{Agrawal01}. 
Iyengar suggested using a classification metric so as to find the optimum generalization. 
Then, a C4.5 decision tree classifier was trained from the optimally generalized training 
data \cite{Iyengar02}. Dowd et al. studied C4.5 decision tree learning from training data 
perturbed by random substitutions. A matrix based distribution reconstruction algorithm 
was applied on the perturbed training data from which an accurate C4.5 decision tree classifier
was learned \cite{Dowd2005}. Inan et al. proposed support vector machine classifiers using 
anonymized training data that satisfy $k$-anonymity. Taylor approximation was used to 
estimate the linear and RBF kernel computation from generalized data\cite{kAnonSvm}. 
Rubinstein et al. studies the kernels of support vector machine in the differential privacy 
and show the trade-off between privacy level and the data utility. They analyze finite 
and infinite dimensional kernels in function of the approximation error under differential 
privacy \cite{RubinsteinLearning09}. Lin at al. studies training support vector classification 
for outsourced data. Random transformation is applied on the training set so that the cloud 
server computes the accurate model without knowing what the actual values are \cite{Lin10}.
Jain et al. studies the support vector machine kernels in the differential privacy setting. 
They propose differentially private mechanisms to train support
vector machines for
interactive, semi-interactive and non-interactive learning scenarios, providing theoretical 
analysis of the proposed approaches \cite{JainKernels13}. 

None of the earlier work has provided a linear classifier directly applicable to anatomized
training data. Such a classifier requires specific theoretical 
and experimental analysis, because anatomized training data provides additional detail 
that has the potential to improve learning; but also additional uncertainty that must be 
dealt with. Furthermore, most of the previous work didn't justify theoretically why the 
proposed heuristics let classifiers generalize well. Therefore, this paper studies 
the following problem: \emph{Define a heuristic to train SVCs and SVMs on anatomized 
data without violating $\ell$-diversity while using the sensitive information,
with a theoretical guarantee of good generalization under reasonable assumptions.}
%
%
%

\section{Definitions and Notations}

\newtheorem{definition}{Definition}

The first four definitions restate standard definitions of 
unprotected data and attribute types.

\begin{definition}
	A dataset ${D}$ is called a \emph{person specific dataset} for population ${P}$ 
	if each instance ${X_i \in D}$ belongs to a unique individual ${p \in P}$.
\end{definition}

The person specific dataset will be called the original training data in this paper.
Next, we will give the first type of attributes.

\begin{definition}
\label{defn:id}
	A set of attributes are called \emph{direct identifying attributes} if they let 
	an adversary associate an instance $X_i \in D$ to a unique individual $p \in P$ 
	without any background knowledge.
\end{definition}

\begin{definition}
\label{defn:qid}
	A set of attributes are called \emph{quasi-identifying attributes}  if there is
	background knowledge available to the adversary that associates
	the quasi-identifying attributes with a unique individual $p \in P$.
\end{definition}

We include both direct and quasi-identifying attributes under the name identifying
attribute. First name, last name and social security number (SSN) are common examples 
of direct identifying attributes.
Some common examples of quasi-identifying attributes are age, postal code,
and occupation. Next, we will give the second type of attribute.

\begin{definition}
\label{defn:s}
	An attribute of instance $X_i \in D$ is called a \emph{sensitive attribute} if we should protect against adversaries correctly inferring the value for
	an individual.
\end{definition}

Patient disease and individual income are common examples of sensitive attributes. 
Unique individuals $p \in P$ typically don't want these sensitive information to be revealed to individuals without a direct need to know that information. 
Provided an instance $X_i \in D$, 
the \emph{class label} is denoted by $X_i.C$. We don't consider the case where $C$ is 
sensitive, as this would make the purpose of classification to violate privacy. $C$ 
is neither sensitive nor identifying in this paper, although our analysis holds for $C$ being an
identifying attribute.

Given the former definitions, we will next define the anonymized training data following 
the definition of $k$-anonymity \cite{kanon_sweeney}.

\begin{definition}
	\label{def:kanondata}
	A training dataset $D$ that satisfies the following conditions is 
	said to be \emph{anonymized training data} $D_k$ \cite{kanon_sweeney}:
	\begin{enumerate}
		\item The training data $D_k$ does not contain any unique identifying 
		attributes.
		
		\item Every instance $X_i \in D_k$ is indistinguishable from at least 
		$(k-1)$ other instances in $D_k$ with respect to its quasi-identifying 
		attributes.
	\end{enumerate}
\end{definition}
Anatomy satisfies a slightly weaker definition; the indistinguishability applies
only to sensitive data.  This will be captured in Definitions \ref{defn:group}-\ref{defn:AnatSVM}.

In this paper, we assume that the anonymized training data $D_k$ is created according to 
a \emph{generalization} based data publishing method. We next define the 
\emph{comparison classifiers}.

\begin{definition}
	A linear support vector classifier (SVC) that is trained on the anonymized training data 
	$D_k$ is called \emph{the anonymized SVC}. Similarly, a support
	vector machine (SVM)  that is trained on the anonymized training data $D_k$ is called 
	\emph{the anonymized SVM}.
\end{definition}

\begin{definition}
	A linear support vector classifier (SVC) that is trained on the original training data 
	$D$ is called \emph{the original SVC}. Similarly, a support vector machine (SVM)  
	that is trained on the original training data $D$ is called \emph{the original SVM}.
\end{definition}

The theoretical aspects of comparison classifiers are out of the scope of this paper.
We will remind the theoretical analysis of the original SVC and SVM classifiers in the
end of this section \cite{Vapnik98}.

We go further from Definition \ref{def:kanondata}, requiring that there must be 
multiple possible sensitive values that could be linked to an individual. 
\textbf{\emph{The proposed
algorithms will be centered around the following definitions}}. This new requirement 
uses the definition of \emph{groups} \cite{ldiversity}.

\begin{definition}\label{defn:group}
\label{defn:gr}
	A \emph{group} ${G_j}$ is a subset of instances in original training data ${D}$ such that 
	${D=\cup_{j=1}^{m} G_j}$, and for any pair ${(G_{j_1},G_{j_2})}$ 
	where ${1 \leq j_1 \neq j_2 \leq m}$, ${G_{j_1} \cap G_{j_2}= \emptyset }$.
\end{definition}

Next, we define the concept of $\ell$-diversity or $\ell$-diverse (multiple possible
sensitive values) for all the groups in the original training data $D$.

\begin{definition}
\label{defn:ldiverse}
	A set of groups is said to be \emph{$\ell$-diverse} if and only if for all groups ${G_j}$ 
	${\forall v \in \Pi_{A_s } (G_j), \frac{freq(v,G_j )} {\vert G_j \vert} \leq \frac{1}{\ell}}$
	 where ${A_s}$ 
	is the sensitive attribute in ${D}$, $\Pi_{A_s }(*)$ is the database $A_s$ projection 
	operation on original training data $*$ (or on data table in the database community),
	${freq(v,G_j )}$ 
	is the frequency of ${v}$ in ${G_j}$ and ${|G_j |}$ is the number 	of instances in
	${G_j}$.
\end{definition}

We extend the data publishing method \emph{anatomization} that is originally 
based on $\ell$-diverse groups by Xiao et al. \cite{XiaoAnatomy}. 

\begin{definition}
	\label{def:anat}
	Given an original training data ${D}$  partitioned in $m$ $\ell$-diverse groups according to
	Definition \ref{defn:ldiverse}, \emph{anatomization} produces an \emph{identifying table}
	 ${IT}$ and a \emph{sensitive table} ${ST}$ as follows. ${IT}$ has schema
	\begin{equation*}
		(C,A_1,...,A_d,GID)
	\end{equation*}
	including the class attribute, the quasi-identifying attributes ${A_i \in IT}$ for 
	${1 \leq i \leq d}$, and the \emph{group id} ${GID}$ of the group $G_j$. For
	each group ${G_j \in D}$ and each instance ${X_i \in G_j}$, ${IT}$ has an instance 
	$X_i$ of the form:
	\begin{equation*}
		(X_i.C, X_i.A_1,...,X_i.A_d,j)
	\end{equation*}
	$ST$ has schema
	\begin{equation*}
		(GID,A_s)
	\end{equation*}
	where ${A_s}$ is the sensitive attribute in ${D}$ and ${GID}$ is the group id of the
	group $G_j$. For each group ${G_j \in D}$ and each instance ${X_i \in G_j}$, ${ST}$ 
	has an instance of the form:
	\begin{equation*}
		(j,X_i.A_s)
	\end{equation*}
\end{definition}

The $IT$ table includes only the quasi-identifying and class attributes. We assume that
direct identifying attributes are removed before creating the $IT$ and $ST$ tables.
We have the following observation from Definition \ref{def:anat} to train a classifier:
\emph{every instance $X_{i} \in IT$ can be matched to $\ell$ instances $X_j \in ST$ using 
the common attribute $GID$ in both data table}. This observation yields 
the \emph{anatomized training data}.

\begin{definition}
	\label{def:anatr}
	Given two data tables $IT$ and $ST$ resulting from the anatomization on original
	training data $D$, the \emph{anatomized training data} $D_A$ is 
	\begin{equation*}
		D_A=\Pi_{IT.A_1, \cdots IT.A_d,ST.A_s} ( \; IT \Join \; ST)
	\end{equation*}
	where $\Join$ is the database inner join operation with respect to the
	condition $IT.GID = ST.GID$ and $\Pi(*)$ is the database projection operation on 
	training data *.
\end{definition}

Anatomized training data shows one of the most na\"{\i}ve data preprocessing approaches.
Another one is ignoring the sensitive attribute in $ST$ table.

\begin{definition}
	\label{def:idtr}
	Given two data tables $IT$ and $ST$ resulting from the anatomization on original
	training data $D$, the \emph{identifying training data} $D_{id}$ is 
	\begin{equation*}
		D_{id}=\Pi_{IT.A_1, \cdots IT.A_d} (\, IT \,)
	\end{equation*}
	where $\Pi(*)$ is the database projection operation on training data *.
\end{definition}

The na\"{i}ve training method of Defintion \ref{def:anatr} is both
costly (a factor of $\ell$ increase in size) and noisy:  for every
true instance, there are $\ell-1$ incorrect instances that may not
be linearly separable.
Ignoring the sensitive data, on the other hand, does not use all the information available in the 
published data (and would likely lead to users insisting on having the original data.) A smarter preprocessing algorithm would eliminate $\ell-1$ instances within 
each group such that the training data becomes separable having good generalization (with or 
without soft margin). This gives the definition of our proposition: \emph{pruned training data}.

\begin{definition}
	\label{def:pruntr}
	Given two data tables $IT$ and $ST$ resulting from anatomization of the original 
	training data $D$, the \emph{pruned training data} $D_P$ is 
	\begin{equation*}
		D_P=\Pi_{IT.A_1, \cdots IT.A_d,ST.A_s} ( \sigma_G ( \; IT ,\; ST))
	\end{equation*}
	where $\sigma_G (IT,ST)$ is a pruning mechanism eliminating $\ell-1$ instances 
	for all groups $G$ in the $IT$/$ST$ pair that are unlikely to be separable (cf. Section \ref{sec:algo}), and
	$\Pi(*)$ is the database projection operation on training data *.
\end{definition}

\begin{definition}
	A linear support vector classifier (SVC) that is trained on the identifying training 
	data $D_{id}$ is called \emph{the identifying SVC}. Similarly, a support vector machine (SVM)
	that is trained on the identifying training data $D_{id}$ is called \emph{the identifying SVM}.
\end{definition}

\begin{definition}\label{defn:AnatSVM}
	A linear support vector classifier (SVC) that is trained on the pruned training 
	data $D_P$ is called \emph{the pruned SVC}. Similarly, a support vector machine (SVM)
	that is trained on the pruned training data $D_P$ is called \emph{the pruned SVM}.
\end{definition}

Now, we are giving the notations of this paper. $X_i$ will denote a training instance in
the original training data $D$ and pruned training data $D_P$ interchangeably. $N$ will be the
total number of instances in $D$ and $D_P$.
$\mathbf{X}$ will be a random variable vector in $D$ 
and $D_P$ interchangeably. $D \subset \mathbb{R}^{d+1}$  and
$D_P \subset \mathbb{R}^{d+1}$ will hold in Euclidean space (see Appendix for practical issues). 
$y$ will be the binary class label with values $\{ -1, 1\}$. 
$f(\mathbf{X})=w\mathbf{X}+b$ will be a linear classifier such that 
$w \in \mathbb{R}^{d+1}$ and $ b \in \mathbb{R}$.
$\mathcal{F}$ is the functional space 
\begin{equation}
	\label{eq:funcsp}
	\{ f: \mathbb{R}^{d+1} \rightarrow \{0,1\}: f(\mathbf{X})=w\mathbf{X}+b,  b \in \mathbb{R}, 
	w \in \mathbb{R}^{d+1}\}.
\end{equation}
We will use $f$ instead of $f(\mathbf{X})$ for shorthand in subsequent parts of this paper. 
The risk of a linear classifier $f$ is $R(f)$ in \eqref{eq:risk}.
\begin{equation}
	\label{eq:risk}
	R(f)=\int \vert (y- f(\mathbf{X}))\vert p(\mathbf{X},y) d\mathbf{X} dy
\end{equation}
In \eqref{eq:risk}, $p(\mathbf{X},y)$ is the joint probability density of training instances 
$\mathbf{X}$ with class label $y$. The empirical risk of classifier $f$ is $\widehat{R}_N(f)$ 
in \eqref{eq:empr}.
\begin{equation}
	\label{eq:empr}
	\widehat{R}_{N}(f)=\frac{1}{N} \overset{N}{\underset{i=1}{\sum}} I(y \neq f(X_i))
\end{equation}
In \eqref{eq:empr}, $N$ is the number of training instances and $I(*)$ is the indicator function. 
\begin{figure*}[t!]
	\centering
	 \begin{subfigure}[b]{0.45\textwidth}
		\centering
		\includegraphics[width=0.6\textwidth,height=0.45\textwidth]{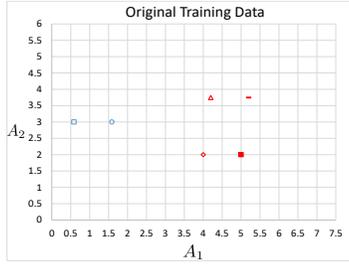}
		\caption{Original Training Data}
		\label{fig:exorg}
	\end{subfigure}
	 \begin{subfigure}[b]{0.45\textwidth}
		\centering
		\includegraphics[width=0.6\textwidth,height=0.45\textwidth]{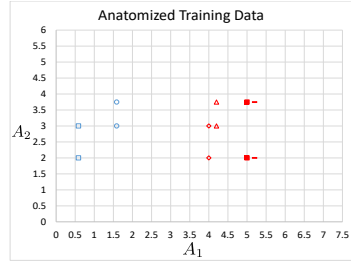}
		\caption{Noisy Training Data ($A_{id}=A_1$, $A_S=A_2$, $\ell=2$)}
		\label{fig:exana}
	\end{subfigure}
	\caption{Toy Example of Training Data with Two Attributes $A_1$ and $A_2$ }.
	\label{fig:ex}
\end{figure*}
The linear classifier $f$ is an empirical risk minimizer such that 
$\widehat{f}_{N}=\underset{f \in \mathcal{F}} {argmin} \widehat{R}_{N}(f)$. Given the empirical
risk minimizer $\widehat{f}_N$ is the SVC with the largest margin, bound \eqref{eq:svbound}
holds
\begin{equation}
	\label{eq:svbound}
	E[R(\widehat{f}_N)] \leq \frac{E[(R \vert \vert w \vert \vert)^2]}{N}
\end{equation}
when the training data is linearly separable \cite{Burges98}. In \eqref{eq:svbound}, $R$ stands for the
radius of the sphere that the shatterable instances lie on and $w$ stands for the weight
vector of hyperplane $f(\mathbf{X})$ in \eqref{eq:funcsp}.
For the same SVC, the generalization
ability is defined in \eqref{eq:genorg} according to VC theory \cite{Vapnik98, Burges98}.
\begin{equation}
	\label{eq:genorg}
	E[R(\widehat{f}_N)] - \underset{f \in \mathcal{F}} {inf} R(f) \leq
		4 \sqrt{\frac{(d+2) log(N+1) + log2}{N}}
\end{equation}
In \eqref{eq:genorg}, $\underset{f \in \mathcal{F}}{inf} R(f)$ is the minimum possible risk for the SVC
$f$. Next, we define our pruning mechanism for the anatomization.

\section{Pruning Mechanism for Anatomization}
\label{sec:algo}

\subsection{Algorithm}
\label{subsec:mot}

We will explain our algorithm ($\sigma_G$ in Definition \ref{def:pruntr}) through the example in Figure \ref{fig:ex}.
The curious reader should visit Figures B.1 and B.2 in the appendix to see the
pseudo code and the complexity. Although the example is for any linear classifier (hyperplane), 
the pruning mechanism is valid for SVC and SVM. We later define the generalization ability of
pruned SVC/SVM (cf. Definition \ref{defn:AnatSVM}).


Figure \ref{fig:exorg} shows the original training data with six instances:  two instances of a blue class 
(on the left side) and four of a red class (on the right side), with two attributes $A_1$ and $A_2$. Here, 
every instance has a different shape and filling combination since they are unique. 
Figure \ref{fig:exana} shows the anatomized training data with 12 instances created from pairs
$IT(A_1,GID)$ and $ST(GID,A_2)$ when $\ell=2$ (cf. Definitions \ref{def:anat}  and \ref{def:anatr}).

\begin{figure}[htb]
	\centering
	\includegraphics[width=0.65\linewidth,height=0.5\linewidth]{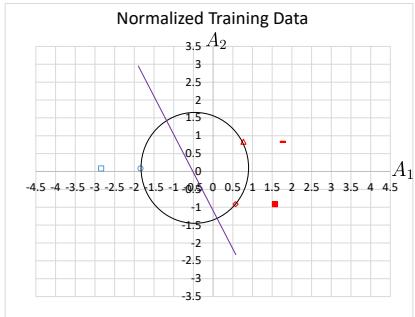}
	\caption{Linear Classifier from Original Training Data}
	\label{fig:orglin}
\end{figure}

A typical training procedure would be the subtraction of mean from attributes $A_1$ and $A_2$ 
in the original training data, and solving an objective function of a perceptron or SVC (cf. Figure \ref{fig:orglin}). 
In Figure \ref{fig:orglin}, the original training data is linearly separable and the instances 
which are closest to the separating hyperplane lie on the surface of the circle 
	\footnote{The discussion can be generalized to sphere for 3 or larger dimensions. 
	See Burges \cite{Burges98} and Vapnik \cite{Vapnik98} for general discussion.}. 
This circle is the key point of linear classification, because the original training data
is guaranteed to be linearly separable if the instances that are closest to the decision boundary
lie on the surface of a circle \cite{Burges98}. 
This observation let us define two steps of the pruning mechanism algorithm:

\begin{enumerate}
	\item \textbf{Prerequisite Step}: Estimate the circle of shatterable instances from the 
	anatomized training data (Algorithm in Figure B.1).
	
	\item \textbf{Pruning Step}: For every group in the anatomized training 
	data, pick an instance that is closest to the surface of the estimated circle of 
	shatterable instances (Algorithm in Figure B.2).
\end{enumerate}

\begin{figure}[htb]
	\centering
	\includegraphics[width=0.65\linewidth,height=0.5\linewidth]{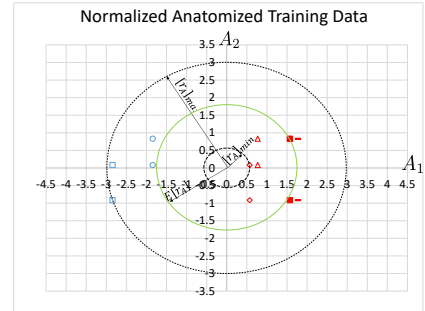}
	\caption{Possible Circles of Shatterable Points}
	\label{fig:possibility}
\end{figure}

Figure \ref{fig:possibility} show the range of radiuses for all possible circles of shatterable instances
in the prerequisite step. The radius of the original training data must be between the norms of the pair of
instances that are closest to (${r_A}_{min}$ in Figure \ref{fig:possibility}) and farthest from
(${r_A}_{max}$ in Figure \ref{fig:possibility}), the origin. Under the random worlds assumption \cite{XiaoAnatomy},
the prerequisite step assumes that $({r_A}_{min},{r_A}_{max})$ has uniform distribution and therefore estimates
the expected radius $E[r]$ with $\frac{{r_A}_{min}+{r_A}_{max}}{2}$ (dashed green line in Figure \ref{fig:possibility}).

\begin{figure}[htb]
	\centering
	\includegraphics[width=0.68\linewidth,height=0.5\linewidth]{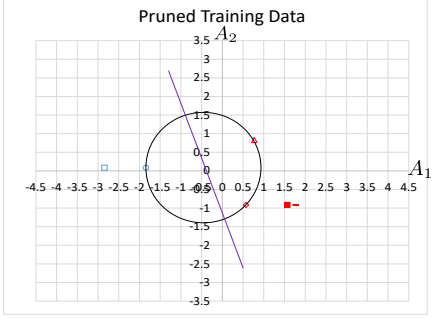}
	\caption{Linear Classifier from Pruned Training Data}
	\label{fig:pruning}
\end{figure}

Using the estimated radius from the prerequisite step, the pruning step creates the pruned
training data in Figure \ref{fig:pruning}. Figure \ref{fig:pruning} also has the hyperplane that
is trained from the pruned training data.
Although the shatterable instances of the pruned training data (cf. Figure \ref{fig:pruning}) are the same as the 
shatterable instances of the original training data (cf. Figure \ref{fig:orglin}), other instances are different. 
The purpose of the pruning step is to find a linearly separable case instead of distribution reconstruction.

There are two remaining issues to address. First is the application of the pruning
algorithm even if the anatomized training data is linearly separable (cf. Figure \ref{fig:exana}). Even though
the anatomized training data is linearly separable in this case, it is not always guaranteed. The
instances within each group are not linearly independent from the other $\ell-1$ instances and the shattering property is 
damaged \cite{Burges98}.
The second issue is non-separable original and anatomized training data. 
If the training data is not linearly separable in the original $(d+1)$ dimensional space, 
the right approach would be projecting it into higher dimensional space, apply the pruning 
algorithm in the projected space and hope for the best with a soft margin classifier.

\subsection{Privacy Preservation}
\label{sssec:priv}

The preprocessing and pruning steps preserve the $\ell$-diversity condition of anatomization. The algorithm
doesn't estimate the correct matchings between the identifying and the sensitive tables. Instead, it 
makes a random guess within each group which is expected to give some linearly separable training data. 
It is possible that the original training data isn't linearly separable or even is a random
set of instances without any pattern (see Section \ref{sec:exps}).

\subsection{Generalization Error of Pruned SVC}
\label{ssec:theory}

	We will now give the upper bound on the generalization error of the pruned SVC (cf. Definition \ref{defn:AnatSVM}). 
	
	\begin{theorem}
		\label{th:generalize}
		Let $N$ be the number of instances, $d$ be the
		number of identifying attributes and $d+1$ be the total number of attributes
		in the original and the pruned training data.
		Let $R$ be the radius of sphere containing the shatterable instances of the
		original training data $D$ and $w$ be the weights of the linear hyperplane
		resulting from linear SV classifier trained on the original training data $D$.
		Let $R_p$ and $w_p$ be
		the symmetric notations for a linear SV classifier trained on the pruned training
		data $D_p$. Assume that all the training instances are located in an Euclidean
		space $\mathbb{R}^{d+1}$. Let $\vert \vert * \vert \vert$ be the Euclidean
		norm of vector $*$. Let $r^2$ be $(R||w||)^2$, $r_p^2$ be 
		$(R_P ||w_p||)^2$, $[r_p^2]_{min}$ be $min\{r_p^2\}>0$ 
		and $[r_p^2]_{max}$ be $max\{r_p^2\} < \infty$. 
		Let $\widehat{R}_{N}(f)$ be the empirical risk of on the original training data $D$ and
		$\widehat{R}_{N_p}(f)$ be the empirical risk on the pruned training data. Let $\mathcal{F}$ 
		be the functional space defining the set of possible linear SV classifiers on the original
		training data $D$ and $\mathcal{F}_p$ be the functional space of possible linear SV 
		classifiers on the pruned training data $D_P$.	
		Let $\widehat{f}_{N}$ be the empirical risk minimizer such that 
		$\widehat{f}_{N}=\underset{f \in \mathcal{F}} {argmin} \widehat{R}_{N}(f)$ and 
		$\widehat{f}_{N_p}$ be the empirical risk minimizer such that 
		$\widehat{f}_{N_p}=\underset{f \in \mathcal{F}_p} {argmin} \widehat{R}_{N_p}(f)$
		Last, let $\underset{f \in \mathcal{F}} {inf} R(f)$ be the lowest value of the risk 
		of the linear SV classifier $f$ that could be analytically calculated. 
		Then, the expected risk $E[R(\widehat{f}_{N_p})]$ of 
		$\widehat{f}_{N_p}$ converges to $\underset{f \in \mathcal{F}} {inf} R(f)$ 
		under the upper bound
		\begin{equation}
			\begin{split}
				\label{eq:anabound}
				E[R(\widehat{f}_{N_p})] - \underset{f \in \mathcal{F}} {inf} R(f) 
				\leq 4 &\sqrt{\frac{(d+2) log(N+1) + log2}{N}}\\
				&+ \frac{[r_p^2]_{max} -[r_p^2]_{min}}{N} 
			\end{split}
		\end{equation}
		using only $D_P$.
	\end{theorem}
	
	The proof of Theorem \ref{th:generalize} is provided in Appendix Section A. The
	upper bound \eqref{eq:anabound} is defined as the function of two terms where the second term is the
	result of using pruned training data. 
%
The former upper bound shows that
pruned SVC can be as accurate as the original SVC under two conditions: \textbf{1) Very large training 
data size ($N \to \infty$) 2) Small size of sensitive attribute domain or low $\ell$ value or both 
($[r_p^2]_{max} -[r_p^2]_{min} \to 0$)}.
%
%
%

Theorem \ref{th:generalize} holds when the pruned training data is mapped into a higher dimensional
space $d^\prime$ using kernel trick. Although the generalization ability of SVMs with RBF kernel is not formally 
defined (invalid Theorem \ref{th:generalize}), SVMs with RBF kernel are expected to work under the 
conditions of Theorem \ref{th:generalize} in the infinite dimensional space \cite{Vapnik98, Burges98}. 

\section{Experiments}
\label{sec:exps}

\subsection{Prerequisites}

\subsubsection{Datasets}

We tested our algorithm on the adult, IPUMS and marketing datasets of the UCI 
data repository \cite{UCIrepository} and the fatality dataset of Keel data repository \cite{keel10}:

\begin{enumerate}
	\item \textbf{Adult:} Adult dataset is drawn from 1994 census data of the United States \cite{UCIrepository}.
	It is composed of 45222 instances after the removal of instances with missing values. 
	The binary classification task is to predict whether a person's adjusted gross income is $\leq50K$ or 
	$>50K$. The attribute ``final weight'' is ignored. Last, education was treated as sensitive attribute
	in the experiments.
	
	\item \textbf{IPUMS:} This data is drawn from the 1970, 1980 and 1990 census data of the
	Los Angeles and Long Beach areas \cite{UCIrepository}. It has 233584 instances in total. 
	We picked the 10 attributes that are included in the adult data.
	The binary classification task is to predict whether a person's total income is $\leq50K$ or $>50K$.
	The classifiers are expected to show a different behavior from the former adult data
	since the population (and to some extent, classification task, as it is total income 
	rather than adjusted gross income) are different. Last, education was treated as sensitive attribute
	in the experiments.
	
	\begin{figure}
		\centering
		\includegraphics[width=0.8\linewidth,height=0.65\linewidth]{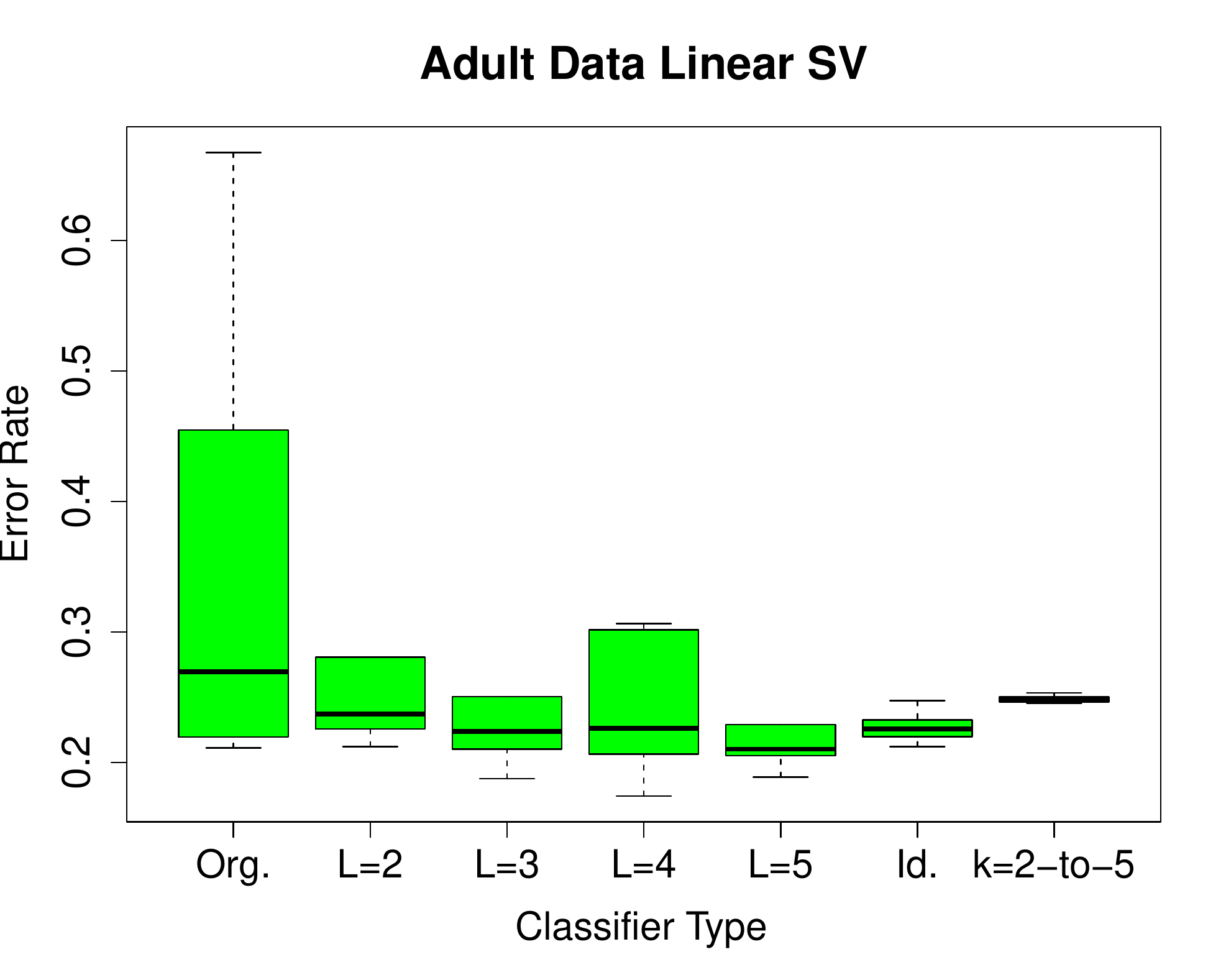}
		\caption{SVC on Adult}
		\label{fig:mainsvcadult}
        \end{figure}  
\begin{figure}  
		\centering
		\includegraphics[width=0.8\linewidth,height=0.65\linewidth]{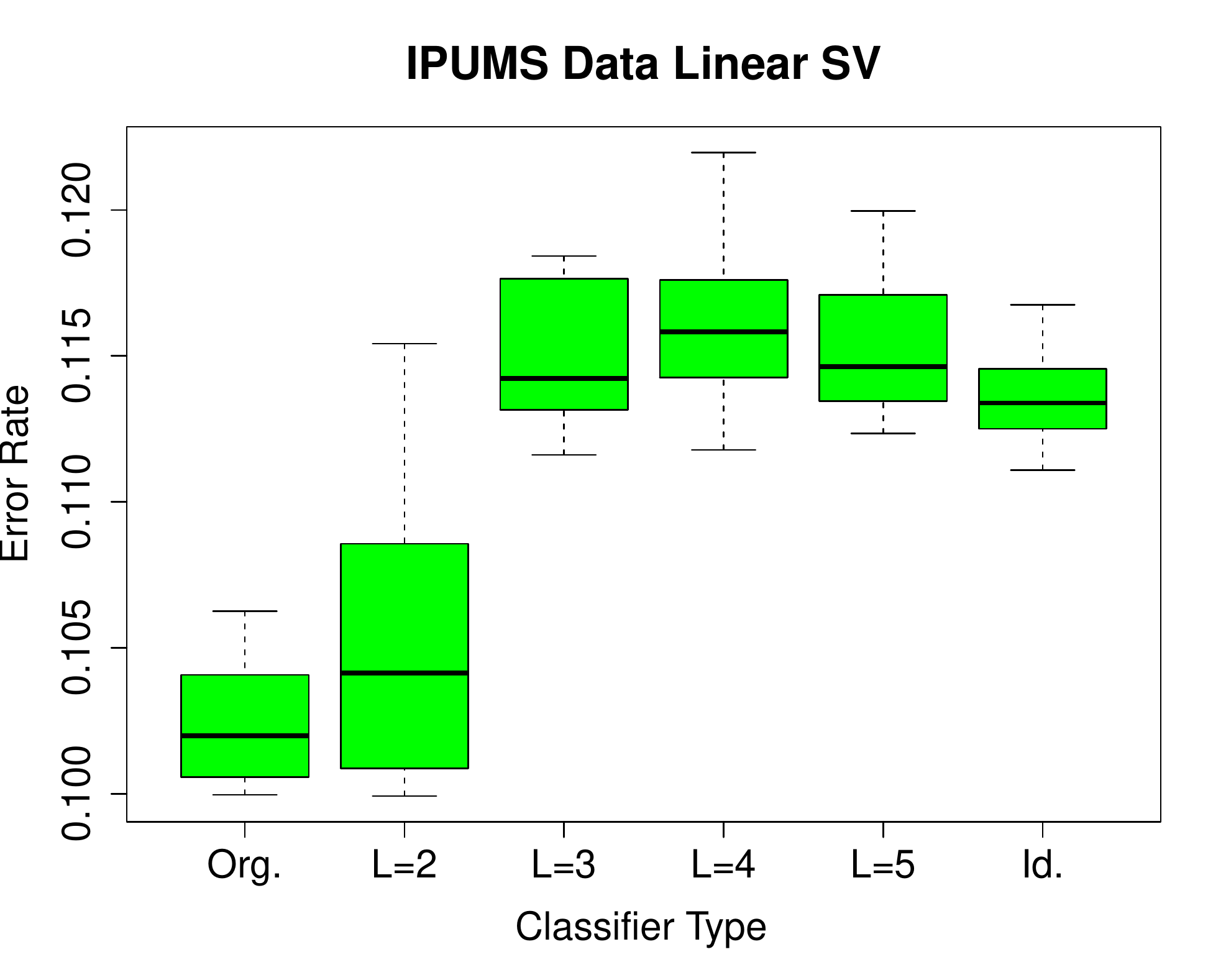}
		\caption{SVC on  IPUMS}
		\label{fig:mainsvcipums}
\end{figure}
\end{enumerate}

The additional information is provided in the appendix for marketing and fatality datasets. Weka was
 used for attribute selection and discretization if needed \cite{Weka}.

\subsubsection{Privacy Setup}

The anatomization was done according to Xiao et al.'s bucketization algorithm \cite{XiaoAnatomy}. 
When $\ell$-diversity condition is not satisfied, the instances were divided into groups of size $\ell$ according to the original bucketization algorithm. Leftover instances were suppressed (not used in
training models). 

Anonymized training data was created for the adult dataset. We used Inan et al.'s value 
generalization hierarchies in the experiments. The privacy parameters were $k=\ell$ for
$k$-anonymity and $\ell$-diversity to compare the classifiers using same group sizes in
training data. 

Anonymized and anatomized training data had the same identifying and sensitive attributes.
The sensitive attributes were chosen such that the $\ell$-diversity is satisfied for at least $\ell=2$.

\subsubsection{Model Evaluation Setup}

LibSVM version 3.21 was used for the support vector classification \cite{Libsvm11}. We will train
the support vector machine with linear (SVC) and RBF kernels (SVM). 

10-fold cross validation was used for evaluation. 
The comparison includes
pruned SVC/SVM, original SVC/SVM and identifying SVC/SVM. The comparison on adult
dataset also include anonymized SVC/SVM. The anonymized SVC/SVM are not included for
other datasets since Inan et al. provided generalization hierarchies only in the adult
dataset \cite{kAnonSvm}. Last, the error rates of pruned and 
original SVC/SVM are compared using the Student $t$-test (See Appendix). Other
models are not included, because Theorem \ref{th:generalize} covers 
only pruned and original SVC/SVM.

\subsection{Analysis of Results}
\label{ssec:expres}

Figures \ref{fig:mainsvcadult}, \ref{fig:mainsvcipums} (see above) and C.1 through 
C.8 (see Appendix) show the boxplots of error rates for SVC and SVM. 
In all Figures, ``Org.''  and ``Id.'' labels will stand for original SVC/SVM and identifying 
SVC/SVM respectively. The pruned and anonymized
SVC/SVM will be represented by their respective privacy parameters (L for $\ell$ and k for $k$.)
This section will 
include the discussion of results in Figures \ref{fig:mainsvcadult} and \ref{fig:mainsvcipums}.
(See Appendix for analysis of other results.)
The analysis have three observation aspects.

The first is the comparison between the pruned and the original SVC/SVM. From 
Theorem \ref{th:generalize}, we expect that the average error rate of the pruned SVC/SVM 
will be higher than the average error rate of the original SVC/SVM because bound \eqref{eq:anabound} 
has the additional second term at the right-hand side (cf. Section \ref{ssec:theory}). Increasing the 
$\ell$ parameter would result in the increase of the average error rate if the training data size
remains same between multiple $\ell$ values (no suppression). 
The rate of the error rate increase in function of $\ell$ is theoretically hard to 
estimate since the assignments of sensitive attributes to each group will be random 
throughout the bucketization algorithm \cite{XiaoAnatomy}. 
Figure \ref{fig:mainsvcipums} show the expected behaviour for the pruned SVC.
The average error rates in Figure \ref{fig:mainsvcadult} show a surprising result.
We observe that the pruned  SVC outperforms the original SVC for multiple $\ell$ values. 
Moreover, the average error rate decreases when $\ell$ is increased. 
Here, the assumption of Theorem \ref{th:generalize} is violated
because of suppression for $\ell \geq 4$. The bound \ref{eq:anabound}
thus does not hold and the result is statistically insignificant (see Appendix.)

The second aspect is the comparison between pruned and identifying SVC/SVM.
In the first case, the identifying SVC/SVM are likely to outperform the pruned SVC/SVM 
if the sensitive attribute is a bad predictor of the class attribute in the original training 
data. The sensitive attribute damages the shattering property of the instances in 
the original training data and the instances near the decision boundary are not on the 
surface of a sphere. The pruned SVC/SVM estimates a model of the original training data 
that is not likely to generalize well. 
Figure \ref{fig:mainsvcadult} shows a bad predictor case, because the average error rates of the identifying SVC is 
less than the average error rates of the pruned SVC when $\ell$ is 2 to 4. $\ell=5$ is a special 
case where the pruning algorithm and $\ell$-diversity show the regularization effect to 
reduce the bias of the underfitting original SVC. In the second case, the pruned SVC/SVM are 
likely to outperform the identifying SVC/SVM if the sensitive attribute is a good predictor 
of the class attribute in the original training data. The symmetric shattering argument implicitly
holds here.
Figure \ref{fig:mainsvcipums} 
shows a special case of a good predictor. The pruned SVC outperforms the identifying SVC only 
for $\ell=2$. For $\ell$ values 3 to 5, the pruning algorithm and $\ell$-diversity act 
like a poorly tuned regularizer that cause overfitting. Unfortunately, we cannot know whether 
the sensitive attribute is a good or bad predictor. Knowing such a behavior would indicate 
the prediction of the sensitive attribute, a defacto violation of $\ell$-diversity. 

The third aspect is the comparison between the pruned and the anonymized SVC/SVM. 
Figure \ref{fig:mainsvcadult} 
show the anonymized SVC/SVM in addition to the anatomized 
and original SVC/SVM. The anatomized SVC/SVM are expected to outperform the
anonymized SVC/SVM because anatomization preserves the original values
for all the attributes. The generalization based $k$-anonymity, on the other hand, 
distorts most of the original attribute values \cite{kAnonSvm}. 
In Figure \ref{fig:mainsvcadult}, the average error rate of the pruned SVC is less than 
the anonymized SVC's when $\ell$ is 3 to 5. These results show the advantage of
anatomization versus generalization-based $k$-anonymity. Anatomization has
high data utility while the sensitive attribute has a strong privacy guarantee, unlike
generalization-based $k$-anonymity.

\section{Conclusion and Future Directions}
\label{sec:conc}

We proposed a preprocessing algorithm for anatomization. Our algorithm
estimates a linearly separable training data from the anatomized training data. 
We defined the generalization ability of support vector classifiers when they are trained
on the former preprocessed data. The key point to remember is that our algorithm gives
good generalization guarantees to support vector classifiers. The proposed mechanism
is evaluated on multiple publicly available datasets and accurate models were observed 
in most cases while $\ell$-diversity is preserved. 

There are multiple future directions for this work. First is the development of other classification 
or clustering algorithms for anatomization. Second is the extension of current work to the 
$k$-anonymity or generalization based $\ell$-diversity. Considering multiple sensitive attributes 
is also another direction.




\section*{Supplementary material}

## B  Pruning Mechanism Algorithm

Figures B.1 and B.2 give the pseudocodes of two steps that are mentioned in Section 4.1.

```
computePrerequisites (D̄_A):
    ‖R_min‖² := +∞  //Squared norm of minimum potential radius
    ‖R_max‖² := −∞  //Squared norm of maximum potential radius
    list_sqNorm := ∅  //List of squared norms
    for i = 1 to |D̄_A|:
        list_sqNorm := list_sqNorm ∪ ‖D̄_A[i]‖²
        if (‖D̄_A[i]‖² ≤ ‖R_min‖²) then
            ‖R_min‖² := ‖D̄_A[i]‖²
        if (‖D̄_A[i]‖² ≥ ‖R_max‖²) then
            ‖R_max‖² := ‖D̄_A[i]‖²
    //Estimate the expected squared radius from U[‖R_min‖², ‖R_max‖²]
    E[‖R‖²] := (‖R_min‖² + ‖R_max‖²) / 2
    return (E[‖R‖²], list_sqNorm)
```

Figure B.1: Prerequisite Step Pseudocode

```
pruneTrainingData (D̄_A):
    (E[‖R‖²], list_sqNorm) := computePrerequisites (D̄_A)
    D_p := ∅  //List holding the pruning result
    i := 0  //index for instances
    j := 0  //index for visited groups
    while (i < |D̄_A|):
        d_c := +∞  //distance between the chosen instance in a group and E[‖R‖²]
        g_c := +∞  //index of the instance in a group with distance d_c
        total_groups := |G_j|  //Total number of instances for all visited groups
        //Look for an instance closest to E[‖R‖²] in the current group
        while (i < total_groups):
            d := |list_sqNorm[i] − E[‖R‖²]|
            if (d < d_c) then  //a closer instance to E[‖R‖²]
                d_c := d
                g_c := i
            i := i + 1
        D_p := D_p ∪ D_A[g_c]
        if (j < |G|) then  //Update number of instances for the next visited group
            j := j + 1
            total_groups := total_groups + |G_j|
    return D_p
```

Figure B.2: Pruning Step Pseudocode

In the pseudocodes, the paramater $\overline{D_A}$ signifies *the augmented anatomized training data*. The augmentation here includes three points:

1. After the inner join operation between the $IT$ and $ST$ tables (cf. Definition 11), the instances are sorted with respect to the attribute $GID$; and then the attributes $IT.GID$ and $ST.GID$ are dropped.

2. The mean of every numeric and non-numeric ordinal attribute is substracted in the augmented anatomized training data. If the attribute $A_i$ is non-numeric ordinal, we replaced the non-numeric values with integer values 1 to $|dom(A_i)|$ according to domain-wise order and set mode of the discrete values to be mean.

3. If the attribute $A_i$ is non-numeric nominal, we replaced the mode of $A_i$ with integer 0 and the rest of $A_i$ values with integer 1.

Note that the pseudocodes use the squared norm instead of the norm itself, because Theorem A.1 defines the generalization error upper bound with the squared radius of sphere containing the shatterable instances of the original training data $D$.

The complexity of the algorithm in Figure B.2 is $O(N(d+2))$. Note that although there are 2 *while* loops in the algorithm, every instance is visited once and the algorithm doesn't go through $d + 1$ attributes due to $list_{sqNorm}$ which makes the execution time of pruning $O(N)$. The prerequisites algorithm need to visit every instance and dimension which makes the execution time $O(N(d+1))$. So the total execution time is $O(N(d+2))$. All the groups ($G$) and the total number of instances within each group ($G_j$) of the anatomized training data are assumed to be known. In case it is not known, the grouping information can be computed using an inner join operation and group by query on $IT$ and $ST$ tables. Such a nested operation is easily implemented in $O(N \log N)$ execution time.

## C  Analysis of Additional Results

**C.1  Additional Datasets** There are 2 more datasets to describe.

**Marketing Data:** This data is drawn from a phone based marketing campaign of a Portuguese banking institution for long term deposits [4]. We created the following binary classification task which is linearly separable under a soft margin SVC: "among all the people who didn't submit a long term deposit, predict whether a person has a housing loan or not". We performed the following preprocessing using Weka filters [30]: 1) pick 39922 instances who didn't make a long term deposit 2) choose four attributes job, day, month and age using the correlation with the class attribute "housing". Discretized age is treated as sensitive attribute.

**Fatality Data:** This data is a U.S. National Center for Statistics and Analysis compilation of 2001 car accidents. The original class attribute has eight labels indicating the level of injury suffered [3]. We created the binary "Injured" and "No_Injury" in the following way: 1) remove the instances with labels "Injured_Severity_Unknown", "Died_Prior_to_Accident", "Unknown" and "Possible_Injury' from the original data. This results in 91085 instances 2) label "Injured" the instances with labels "Nonincapacitating_Evident_Injury", "Incapacitating_Injury" and "Fatal_Injury". No feature selection is

applied on this dataset. "POLICE_REPORTED_ALCOHOL_INVOLVEMENT" was treated as sensitive attribute.

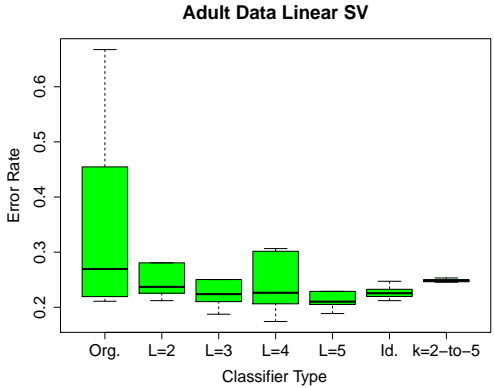

Figure C.1: SVC on Adult

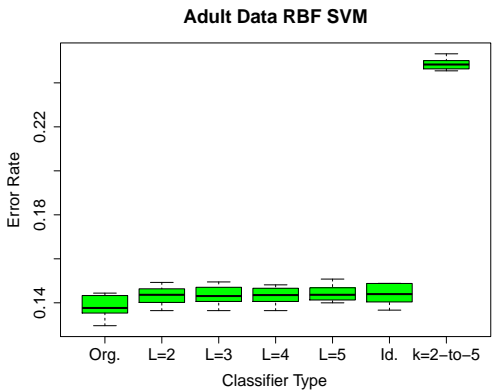

Figure C.2: SVM on Adult

**C.2 Analysis of Additional Results** We will analyze more results in this section. In Section 5, three observation aspects were discussed for evaluation.

To recall, *the first aspect* was the comparison between the pruned and original SVC/SVM. From Theorem A.1, we expect that the average error rates of pruned SVC/SVM will be greater then the original SVC/SVM's if there is no suppression due to $\ell$-diversity constraint. One exceptional case would be the regularization effect where $\ell$-diversity and pruning algorithm reduces either the bias of underfitting SVC/SVM or the variance of overfitting SVC/SVM. Another exceptional case would be the suppression of many instances of the original training data due to $\ell$-diversity constraint. This violates the assumption of Theorem A.1. *The second aspect* is the comparison between pruned and identifying SVC/SVM. From the shattering properties of the statistical learning theory, the pruned SVC/SVM are expected to outperfom the identifying SVC/SVM if the sensitive attribute is a good predictor of the class attribute. If sensitive attribute is, on the other hand, a bad predictor of the class attribute; the opposite of the former behaviour is expected to occur. Last, *the third aspect* is the comparison between the pruned and anonymized SVC/SVM. The pruned SVC/SVM are expected to outperform the anonymized SVC/SVM because anatomization preserves the original values for all the attributes. Figures C.1 to C.8 show the results of all the experiments. In all Figures, "Org." and "Id." labels will stand for the original SVC/SVM and the identifying SVC/SVM respectively. The pruned and anonymized SVC/SVM will be represented by their respective privacy parameters (L for $\ell$ and k for $k$.)

Figure C.1 shows a surprising result of pruned SVC in the first and second aspect. In the first aspect, increasing $\ell$ reduces the average error rate of the pruned SVC so $\ell$-diversity and pruning algorithm regularize the underfitting original SVC. Theorem A.1 does not hold as well for $\ell \geq 4$, because some original training data instances are suppressed. In the second aspect, the sensitive attribute is a bad predictor of the class attribute. Identifying SVC performs better than many pruned SVCs. In the third aspect, the average error rate of pruned SVC is less than the average error rate of anonymized SVC for all $\ell = k$ values (See Section 5.2).

Figure C.2 shows the expected results of pruned SVM in all three aspects. Increasing $\ell$ result in the increase of average error rate for pruned SVM and original SVM outperforms the pruned SVM. The expectation from Theorem A.1 occurs here despite suppression (violation of assumption). We believe that $\ell$-diversity and pruning algorithm act as regularizer for SVMs with RBF kernel which tend to overfit to the training data. Notice that the sensitive attribute is a good predictor of the class attribute in the infinite dimensional space since average error rate of pruned SVM is less than the average error rate of identifying SVM. Last, the average error rate of the pruned SVM is less than the anonymized SVM's one by 0.1.

Figure C.3 shows in general the expected result of pruned SVC in the first aspect. $\ell = 4$ is a special case where its average error rate is greater than the pruned SVC's that is trained on 5-diverse data. Theorem A.1's assumption is violated again when $\ell = 5$ because some original training data instances are suppressed. In the second aspect, the pruned SVC cannot capture the good shattering property that the sensitive attribute provide in the original dimensional space.

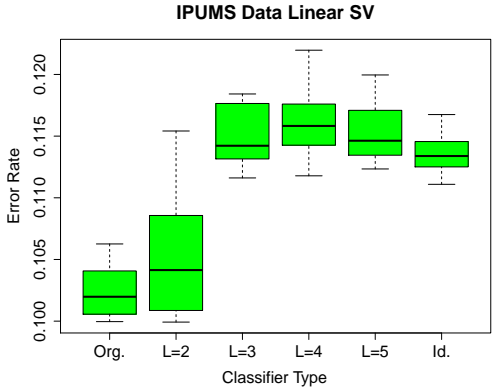

Figure C.3: SVC on IPUMS

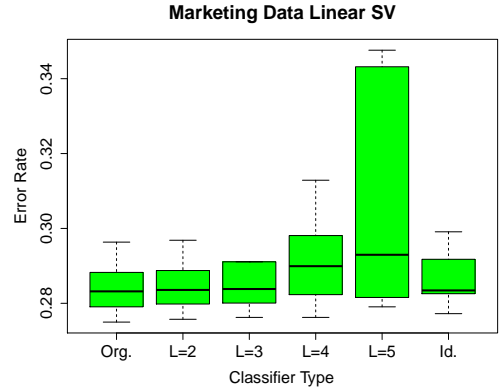

Figure C.5: SVC on Marketing

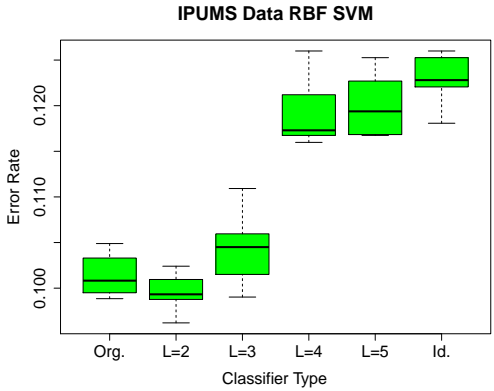

Figure C.4: SVM on IPUMS

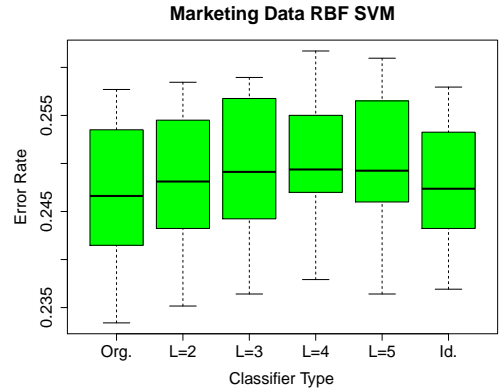

Figure C.6: SVM on Marketing

Figure C.4 shows in general the expected result of pruned SVM for first and second aspects. In the first aspect, the pruned SVM outperforms the original SVM when $\ell = 2$. This shows that the pruning algorithm and $\ell$-diversity has the regularization effect even if the sensitive attribute is a good predictor according to the second aspect. The regularization case could occur in general, because it is statistically significant for confidence interval 0.95 (See Table 2).

Figure C.5 show the expected behaviour of pruned SVC in the first aspect. One thing to emphasize is the surprising spike in the error rate distribution when $\ell = 5$. The reason is that the original training data satisfies the $\ell$-diversity condition when $\ell = 2$ and $\ell = 3$. When $\ell = 5$, almost half of the training instances are suppressed. This strongly violates the assumption of Theorem A.1 and the result is also not statistically significant (cf. Table 1). We should note that the sensitive attribute is a bad predictor since the average error rates of pruned SVC are greater than the identifying SVC's.

Figure C.6 show the expected behaviour of pruned SVC in the first aspect. The pruned SVC also show the expected result in the second aspect. The sensitive attribute is not a good predictor in the infinite dimensional space.

In Figure C.7, the pruned SVC gives an interesting and surprising result in the first aspect. The average error rate of pruned SVC is approximately same as the average error rate of original SVC for all $\ell$ values. We believe that this would only occur in this dataset because the results are not statistically significant (cf. Table 1.) When $\ell = 3$ and $\ell = 4$, the assumption in Theorem A.1 is violated because most of the training instances are suppressed. In the second aspect, sensitive attribute is bad (insignificant) predictor since the pruned SVC thus does not reduce the error rate of the identifying SVC.

Figure C.8 show the expected results of pruned SVM in the first two aspects (despite violating the assumption of Theorem A.1). In the second aspect, note that the sensitive attribute is a bad predictor in

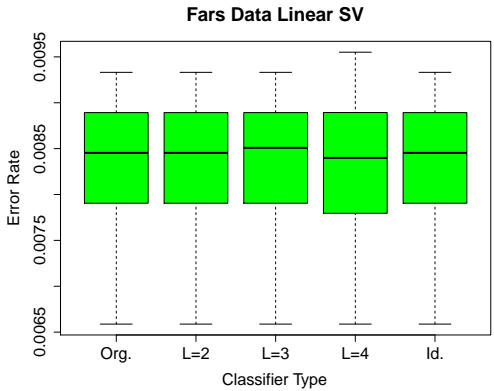

Figure C.7: SVC on Fatality

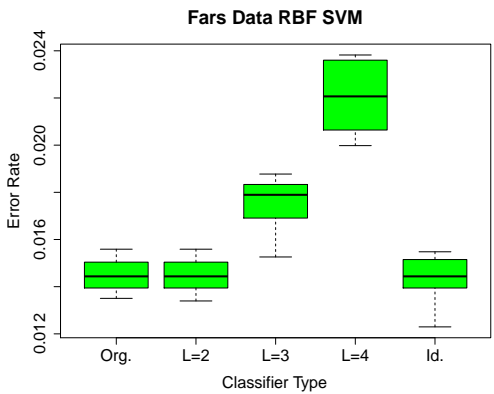

Figure C.8: SVM on Fatality

the inifinite dimensional space. The average error rate of pruned SVM is greater than the identifying SVM's.

### C.3 Student $t$-test for Pruned SVC/SVM versus Original SVC/SVM
Tables 1 and 2 give the statistical test results for *confidence interval 0.95*. In all Tables, "P" stands for pass while "F" stands for fail. "N/A" stands for not applicable in cases where the domain size of sensitive attribute is less than the $\ell$ value. "Org." stand for the original SVC/SVM whereas "$\ell$" stand for the pruned SVC/SVM. Note that we do the test for original SVC/SVM vs pruned SVC/SVM, because the Theorem A.1's scope covers this analysis.

In Section C.2, we saw the theoretically expected results for pruned SVC vs original SVC when they are trained on IPUMS dataset (cf. Figure C.3). Table 1 shows that the difference between the pruned and original SVC is statistically significant for almost all $\ell$ values. We saw, in contrary, theoretically unexpected results in case of pruned SVC vs. original SVC on adult and fatality datasets. Table 1 shows that the difference between

| Dataset | Org. vs $\ell$=2 | Org. vs $\ell$=3 | Org. vs $\ell$=4 | Org. vs $\ell$=5 |
|---|---|---|---|---|
| Marketing | P | F | F | F |
| Fatality | F | F | F | N/A |
| IPUMS | F | P | P | P |
| Adult | F | F | F | F |

Table 1: Pruned SVC vs Original SVC

the pruned and original SVC are statistically insignificant in adult and fatality datasets. So, the theoretically unexpected results are likely to occur by just random chance and it doesn't make Theorem A.1 invalid. In marketing dataset, the difference between the pruned and the original SVC is statistically insignificant for $\ell$ values 3-to-5, because most of the training instances were suppressed. Note that Theorem A.1 is valid if and only if both the pruned and the original training dataset have the same number of instances (no or negligeable suppression.) (See Theorem A.1)

| Dataset | Org. vs $\ell$=2 | Org. vs $\ell$=3 | Org. vs $\ell$=4 | Org. vs $\ell$=5 |
|---|---|---|---|---|
| Marketing | P | P | P | P |
| Fatality | F | P | P | N/A |
| IPUMS | P | P | P | P |
| Adult | P | P | P | P |

Table 2: Pruned SVM vs Original SVM

Table 2 shows that the difference between pruned SVM and original SVM are statistically significant in almost all datasets for multiple $\ell$ values. In Section C.2, the expectation from Theorem A.1 occured in all the datasets. (cf. Figures C.2, C.4, C.6 and C.8) This good result are therefore very unlikely to occur by random chance and Theorem A.1 is valid in infinite dimensions. This result is in parallel to Vapnik and Burges' claim for the original SVM when there is no suppression [29, 5]. Last, we observe surprisingly significant results when the assumption of Theorem A.1 is violated. We believe that the $\ell$-diversity and pruning acts as a regularizer since SVMs with RBF kernel tend to overfit to the training data.

*In summary, we measured the statistically significant error rates when the pruned SVC/SVM show the expectation from Theorem A.1. The error rates were statistically insignificant when they don't respect the expected result of Theorem A.1 or when the pruned training data violates the assumption of Theorem A.1.*



\end{document}